\def\year{2017}\relax
\documentclass[letterpaper]{article}
\usepackage{aaai17}
\usepackage{times}
\usepackage{helvet}
\usepackage{courier}
\usepackage{graphicx}
\usepackage{url}            
\usepackage{color}
\usepackage{multirow}

\begin{document}
%
\title{Leveraging Video Descriptions to Learn Video Question Answering}
\author{
\normalsize
Kuo-Hao Zeng$^*$$^{\dagger}$, 
Tseng-Hung Chen$^*$, 
Ching-Yao Chuang$^*$ 
Yuan-Hong Liao$^*$,
Juan Carlos Niebles$^{\dagger}$, 
Min Sun$^*$ \\ \\
$^*$Department of Electrical Engineering, National Tsing Hua University\\
$^{\dagger}$Department of Computer Science, Stanford University \\
}

\maketitle
\begin{abstract}
We propose a scalable approach to
learn video-based question answering (QA):
to answer a free-form natural language question about
the contents of a video.
Our approach automatically harvests a large number of videos and descriptions freely available online.
Then, a large number of candidate QA pairs are automatically generated from descriptions rather than manually annotated.
Next, we use these candidate QA pairs to train a number of video-based QA methods extended from MN~\cite{NIPS2015_5846}, VQA~\cite{VQA}, SA~\cite{VideoAttRNN}, and SS~\cite{SS15}.
In order to handle non-perfect candidate QA pairs, we propose a self-paced learning procedure to iteratively identify them
and mitigate their effects in training.
Finally, we evaluate performance on manually generated video-based QA pairs.
The results show that our self-paced learning procedure is effective, and
the extended SS model outperforms various baselines.
\end{abstract}


\section{Introduction}\label{sec.intro}

Understanding video contents at human-level is a holy grail in
visual intelligence.
Towards this goal, researchers have studied intermediate tasks such as detection of objects and events,
semantic segmentation, and video summarization.
Recently, there has been increased interest in many tasks that bridge language and vision,
which are aimed at demonstrating abilities closer to human-level understanding.
For example, many researchers \cite{title_generation,xu2016msr,PanMYLR15,PanXYWZ15,YuWHYX15,hendricks2015deep} have worked on 
video captioning and generated natural language descriptions of videos recently.
Despite the great progress, video captioning suffers from similar issues as image captioning:
(1) it is fairly easy to generate a relevant, but non-specific, natural language description \cite{showtell2015};
(2) it is hard to evaluate the quality of the generated open-ended natural language description.

An alternative task that addresses these issues is visual question answering (QA) \cite{VQA}, which brings two
important properties:
(1) specific parts of a visual observation need to be understood to answer a question;
(2) the space of relevant answers for each question is greatly reduced.
Thanks to these properties, visual QA has become a viable alternative towards human-level visual understanding at a finer level of detail.
Moreover, with the reduced answer space, simple metrics such as standard accuracy (percentage of correct answers) can be used to evaluate performance.

\begin{figure}[!t]
\centering
		\includegraphics[width=\linewidth]{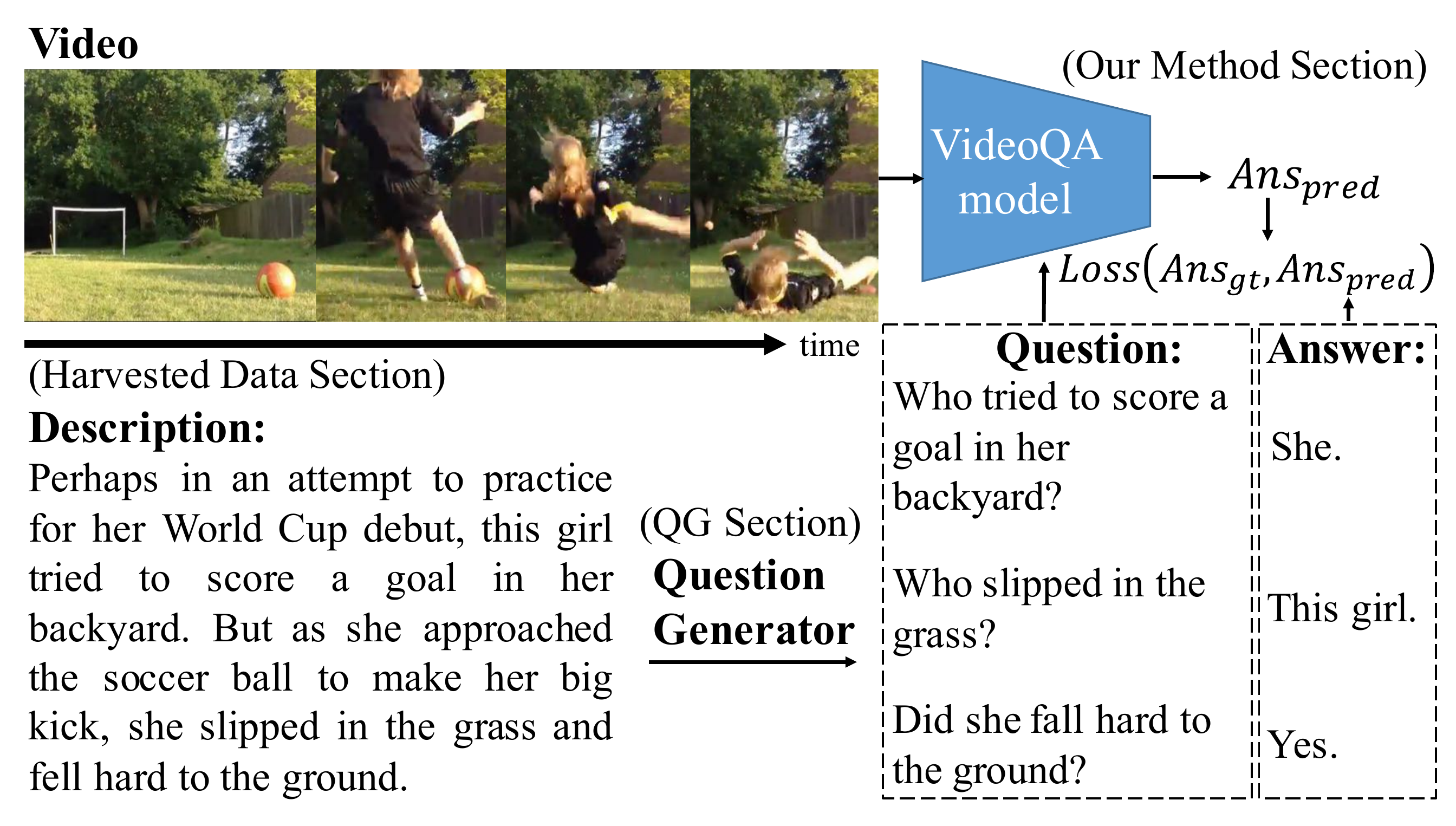}
    	\caption{\small Illustration of our approach. Given harvested videos and descriptions (see harvested data section), our system automatically generate question-answer pairs from descriptions (see questions generation section (QG section)). Then our VideoQA model takes a video and the generated questions as input and outputs the corresponding answers (see our method section). $Ans_{pred}$ denotes the predicted answer and $Ans_{gt}$ denotes the ground truth answer.}\label{fig:teaser}
    	\end{figure}

The biggest drawback of visual QA comes from the significant human efforts required to build benchmarking datasets.
Most current collection techniques \cite{VQA,NIPS2014_5411} require humans to view the visual data and manually create QA pairs for both training and testing.
Furthermore, the situation becomes worse when the data consists of \textit{videos} rather than images.
One of the earliest attempts to create a QA benchmark for videos
is the MovieQA dataset by \cite{MovieQA}.
Since it is expensive to hire annotators to watch entire movies,
plot synopses are used as a proxy during the first step.
Human annotators may form any number and type of questions for each plot paragraph.
Given the initial set of questions, annotators are asked to localize context in the movie to answer the question.
Annotators may correct the questions if they cannot localize context in the movie.
Finally, annotators provide one correct answer and four wrong answers.
In total, MovieQA consists of $14944$ QA pairs from $408$ movies.

The MovieQA dataset and the approach to collect data have the following limitations.
First, it is unknown how to create a large-scale QA dataset with videos in the wild without available plots to be used as a proxy.
Second, the task of picking one correct answer out of five candidate answers is less challenging than the task with $1K$ answer space in VQA~\cite{VQA}.

In this paper, we aim at building a video QA dataset that does not require the manual construction of QA pairs
for training (see Fig.~\ref{fig:teaser} for our workflow).
We propose to leverage the fact that Internet videos with user-curated descriptions
can be easily harvested at a large-scale.
We adopt a state-of-the-art question generation method~\cite{Heilman2010} to generate candidate QA pairs automatically from descriptions.
With this approach, we have collected a large-scale video QA dataset with $18100$ videos and $175076$ candidate QA pairs.

While the automatic generation of QA pairs can scale very well, it is not perfect. In fact, we observe that $10\%$ of the automatically generated pairs are irrelevant/inconsistent to the visual content in the corresponding video.
As we will show, current supervised learning frameworks for video QA can be harmed by non-perfect training QA pairs.
To tackle this challenge, we introduce a novel ratio test to automatically identify non-perfect candidate QA pairs
and a self-paced learning procedure to iteratively train a better model.
Furthermore, we demonstrate that this strategy is widely applicable by extending several existing models that bridge vision and language to tackle the problem of video-based QA.

We extend four methods for our video-based QA task: MN~\cite{NIPS2015_5846}, VQA~\cite{VQA}, SA~\cite{VideoAttRNN}, and SS~\cite{SS15}.
We empirically evaluate their performance on $2000$ videos associated with about $2500$ manually generated ground truth QA pairs.
Our results show that self-paced learning is effective and the extended SS method outperforms other baselines.

\section{Related Work}\label{sec.RW}

\noindent\textbf{Image-QA.} There has been a significant recent interest in
image-based visual question answering \cite{Bigham:2010:VNR,Geman24032015,NIPS2014_5411,Malinowski_2015_ICCV,VQA,gao2015mQA,noh2015dppnet,AndreasCVPR16,MaAAAI16},
where the goal is to answer questions given a \textit{single image} as visual observation.
In the following, we discuss a few of them which have collected their own Image-QA dataset.
\cite{Bigham:2010:VNR} use crowdsourced workers to complete Image-QA task asked by visually-impaired users in near real-time. \cite{Geman24032015,NIPS2014_5411} are pioneers on automatic visual question answering, but only consider question-answer pairs related to a limited number of objects, attributes, etc.
\cite{NIPS2014_5411} also propose a new evaluation metric (WUPS) that accounts for
word-level ambiguities in the answers, which we adopt for our experiments.
\cite{Malinowski_2015_ICCV} further propose a sequence-to-sequence-like model for Image-QA
and extend their previous dataset~\cite{NIPS2014_5411}.
\cite{VQA} manually collected a large-scale free-form and open-ended Image-QA dataset.
They also propose a model which embeds question and image into a joint representation space.
\cite{gao2015mQA} collected a Freestyle Multilingual Image Question Answering (FM-IQA) dataset
consisting of Chinese question-answer pairs and their English translation.
They also propose a sequence-to-sequence-like model with two set of LSTMs: one for questions and one for answers.
Most methods require being trained with manually collected visual QA data, which must be correct.
In contrast, we propose a novel way to harvest and automatically generate our own video QA dataset,
which scales to an enormous number of QA pairs with the cost of potentially containing non-perfect QA pairs.
This creates a challenge to existing methods, for which leveraging our large number of examples is risky due to potentially non-perfect training examples. We tackle this issue by introducing a self-paced learning procedure to handle non-perfect QA pairs during training.

\noindent\textbf{Question generation.}
Automatic question generation is an active research topic by itself.
Most existing question generation methods \cite{QGworkshop09,QGChallenge09,GatesThesis08}
focus on generating questions in specific domains such as English as a Second Language (ESL).
For our purposes, it is important to generate a diverse set of QA pairs that can match the open nature of the user-generated video domain.
In particular, we adopt the method from
\cite{Heilman2010} to generate candidate QA pairs from video description sentences.
Their method consists of a statistical ranking based framework for the generation of QA pairs in open domains.
In a similar spirit, \cite{NIPS2015Ren} propose to automatically generate QA pairs from image description for the image-based QA task.
However, they focus on generating high-quality questions by constraining their
structure to four types of questions: objects, numbers, color, and location-related questions.
In contrast, our goal is to generate an enormous number of open-domain questions that can be used to train data-demanding models such as deep learning models.

\noindent\textbf{Video-QA.}
In contrast to the Image-QA task, video-based QA is a much less explored task.
\cite{TuMLCZ14} have studied joint parsing of videos and corresponding text to answer queries.
\cite{MovieQA} recently collect a Multimodal QA dataset consisting movie clips, plot, subtitle, script, and Described Video Service (DVS).
Similar to most Image-QA datasets, they ask human annotators to generate \textit{multiple choice} QA pairs.
This approach requires an enormous amount of human efforts since annotators must verify that the context of the answer to the question can be localized in the movie.
\cite{zhu2015uncovering} collect a larger video-based QA dataset with $390744$ \textit{fill-in-the-blank} questions
automatically generated from other manually created video caption datasets.
Our proposed method focus on answering \textit{free-form natural language questions} rather than a \textit{fill-in-the-blank} questions.
Moreover, our videos and descriptions are harvested from an online video repository without any additional manual effort to generate descriptions. Hence, we believe our proposed method further advances towards a large-scale setting for the video-based QA task.

\begin{figure*}[!t]
	\centering
    	\includegraphics[width=0.8\textwidth]{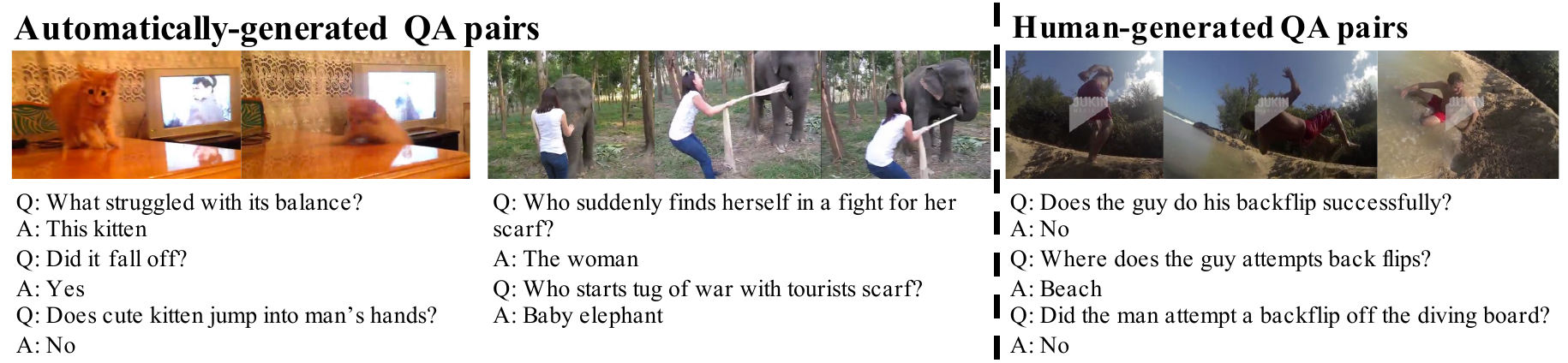}
    	\caption{\small Sample videos and question-answer pairs in our Video-QA dataset. This dataset contains $18100$ open-domain videos, including $151263$ and $21352$ automatically generated question-answer pairs in the training and validation sets (Left-panel) and $2461$ human-generated question-answer pairs in the testing set (Right-panel).}\label{fig:sample}
    	\end{figure*}

\section{Video Question Answering Dataset}\label{sec.vqad}
We describe the harvested data for our new Video Question Answering (Video-QA) dataset.
We start by crawling an online curated video repository ({\footnotesize \url{http://jukinmedia.com/videos}})
to collect videos with high-quality descriptions.

\subsection{Harvested Data\footnote{Available at \url{http://aliensunmin.github.io/project/video-language/}}}\label{sec.HD}

\noindent\textbf{Internet videos.}
We collected $18100$ open-domain videos with average duration of $1.5$ minutes ($45$ seconds median).
Our videos are typically captured from handheld cameras,
so the video quality and amount of camera motion vary widely.

\noindent\textbf{Descriptions.}
Originally, each video is associated with a few description sentences submitted by the video owner.
Then, staff editors of the video repository curate these sentences by removing abnormal ones.
As a result, there are typically 3-5 description sentences for each video,
as shown in Fig.~\ref{fig:teaser}.
The description contains details of the scene (e.g., backyard), actor (e.g., the girl), action (e.g., score), and possibly non-visual information
(e.g., practice for her World Cup debut).

\subsection{Questions Generation (QG)}\label{sec.GQ}

\noindent\textbf{Candidate QA pairs.}
We apply an state-of-the-art question generation method~\cite{Heilman2010} to automatically
generate candidate QA pairs (auto-QG) for each description sentence.
We expect that some candidate QA pairs are not perfect.
In our method section, we will describe our strategy to handle these non-perfect QA pairs.

\noindent\textbf{Generating questions with the answer \emph{No}.}
The state-of-the-art question generation method~\cite{Heilman2010}
can only generate \emph{Yes/No} questions with the answer \emph{Yes}.
In order to obtain a similar number of questions with answer \emph{No},
we use the existing \emph{Yes/No} questions of each video to retrieve similar \emph{Yes/No} questions associated
to other videos. Since the retrieved questions are most likely irrelevant/inconsistent with respect to the video content,
we assign \emph{No} as their answer.
In total, we have $174,775$ candidate QA pairs. Examples are shown in Fig.~\ref{fig:sample}.
Among them, $151062$ QA pairs from $14100$ videos are used for training, and $21252$ QA pairs from $2000$ videos are used for validation.
The remaining $2000$ videos are used for testing.

\noindent\textbf{Verified QA pairs.}
To improve the quality of QA pairs generated by auto-QG, we ask users on Amazon Mechanical Turk
to manually clean the subset of candidate QA pairs in two steps.
First, each turker is given five QA pairs corresponding to one video.
The turker decides whether each QA pair is correct, irrelevant, or can-be-corrected.
We move QA pairs selected as \emph{can-be-corrected}
into the second step, where we ask turkers to correct each QA pair.
Only a small portion (about $10\%$) of QA pairs require the second step.

\noindent\textbf{Human-generated QA pairs.}
To evaluate Video-QA performance, we collect $2461$ human generated QA pairs associated with the testing videos.
First, in-house annotators remove descriptions which are irrelevant to the video content.
Then, we ask Amazon Mechanical Turk (AMT) workers to generate QA pairs according to the titles and descriptions.
This process is time-consuming, similar to the procedure used in MovieQA~\cite{MovieQA}.
To encourage diversity in the QA pairs, each video is assigned to two different workers.
Finally, we keep the QA pairs which have answers within the union set of the answers in training.

\begin{figure}[!t]
	\centering
    	\includegraphics[width=0.47\textwidth]{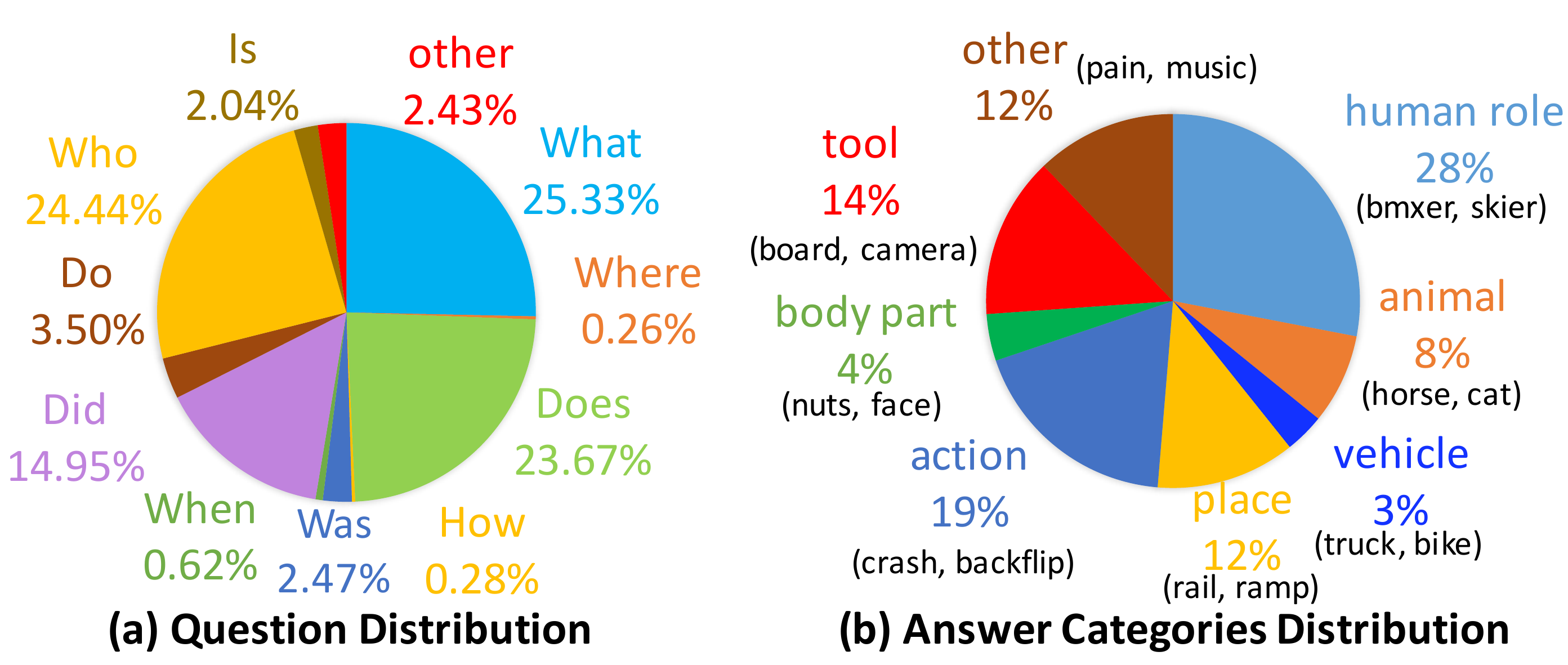}
    \caption{\small Question and answer distribution. \textbf{(a)} Question distribution based on the words that start the question. \textbf{(b)} Answer (\emph{Yes/No} answers excluded) distribution on eight manually defined categories. Two typical answers in each category are also shown.}\label{fig:q_d}
    	\end{figure}

\subsection{Questions and Answers Analysis}

\noindent\textbf{Questions.} We categorize questions based on the words that start the question and show their distribution in Fig.\ref{fig:q_d}(a). Our Video-QA dataset contains diverse questions, including 5W1H questions. Moreover, because our QA task is based on video content, several questions refer to actions or motions. Specifically, the large proportion of auxiliary verbs such as \emph{Does}, \emph{Did}, and \emph{Do} indicate that many of our questions are about the main verbs in the event description.
This shows our questions are quite different compared to the image-based QA datasets~\cite{VQA,NIPS2015Ren,NIPS2014_5411}, which are mainly about objects, colors, and numbers.
The maximum, minimum, mean, standard deviation, and median lengths of our questions are $36$, $2$, $10.8$, $5.3$, $9$, respectively.
See \cite{VTGtech} for more analysis on Human-generated QA pairs and a comparison between Automatically-generated QA pairs and Human-generated QA pairs.

\noindent\textbf{Answers.}
We show the answer (\emph{Yes/No} answers excluded) distribution on eight manually defined categories in Fig.\ref{fig:q_d}(b).
Two typical answers in each category are shown.
Instead of objects, colors, and numbers, as in most Image-QA datasets, our answers contains a large portion of human roles and actions.
Note that \emph{Yes} and \emph{No} account for $32.5\%$ and $32.5\%$ of the whole set, respectively.

\section{Our Method}\label{sec.ourS}

Video-QA consists of predicting an answer $\textbf{a}$ given a question $\textbf{q}$ and video observation $\textbf{v}$.
We define a video as a sequence of image observations $\textbf{v}=\left[ v^1, v^2, \dots\right]$, and both answer and question as a natural language sentence (i.e., a sequence of words) $\textbf{a}=\left[a^1, a^2,\dots\right]$ and $\textbf{q}=\left[q^1, q^2,\dots\right]$, respectively.
To achieve Video-QA, we propose to learn a function $\textbf{a}=f(\textbf{v},\textbf{q})$, where $\textbf{v},\textbf{q}$ are the inputs and $\textbf{a}$ is the desired output.
Given a loss $L(\textbf{a},f(\textbf{v},\textbf{q}))$ measuring the difference between a truth answer $\textbf{a}$ and a predicted answer $f(\textbf{v},\textbf{q})$, we can train function $f(\cdot)$ using a set of $(\textbf{v}_i,\textbf{q}_i,\textbf{a}_i)_i$ triplets (indexed by $i$) automatically generated
from videos and their description sentences.
As mentioned earlier, the automatically generated QA pairs inevitably include some non-perfect pairs which are irrelevant or inconsistent with respect to the video content.
We propose a novel test ratio and a self-pace learning procedure to mitigate the effect of non-perfect QA pairs during training.

\subsection{Mitigating the Effect of Non-perfect QA Pairs}\label{sec.noiseH}
The key to mitigating the effect of non-perfect pairs is to automatically identify them.
We follow our intuition below to design a test to identify non-perfect pairs.
Intuitively, if a training question answer pair is \emph{relevant/consistent} with respect to a video content,
the loss $L(\textbf{a},f(\textbf{v},\textbf{q}))$ should be small.
If we keep the same QA pair, but change the video content to a dummy video $\textbf{v}_D$ with all zero observation,
the loss $L(\textbf{a},f(\textbf{v}_D,\textbf{q}))$ should increase significantly.
In contrast, if another training question answer pair is \emph{irrelevant/inconsistent} with respect to a video content,
the loss $L(\textbf{a},f(\textbf{v},\textbf{q}))$ should be large.
Moreover, if we keep the same QA pair, but change the video content to a dummy video $\textbf{v}_D$,
the loss $L(\textbf{a},f(\textbf{v}_D,\textbf{q}))$ should not change much.
Our intuition suggests that the loss of a non-perfect triplet $(\textbf{v}_i,\textbf{q}_i,\textbf{a}_i)_i$ is less sensitive to the change of video content,
compared to the loss of an ideal triplet.

\noindent\textbf{Ratio test.}
Following the intuition, we calculate the ratio $r$ as the dummy loss $L(\textbf{a},f(\textbf{v}_D,\textbf{q}))$
divided by the original loss $L(\textbf{a},f(\textbf{v},\textbf{q}))$.
If the ratio $r$ is small, it implies the training triplet is non-perfect.

\noindent\textbf{Self-paced learning.}
Firstly, we use all the training triplets to learn a reasonable function $f(\cdot)$.
Once we have the initial function $f(\cdot)$, we can calculate ratio for every training triplet.
For a video with a ratio smaller than a threshold $\gamma$ (i.e., satisfied the ratio test),
we change its training video into the dummy video $\textbf{v}_D$.
Then, we re-train the function $f(\cdot)$.
Given a new function, the same steps can be repetitively applied.
The whole self-paced procedure stops after no addition videos satisfied the ratio test.

\begin{figure*}[!t]
\centering
    	\includegraphics[width=0.8\textwidth]{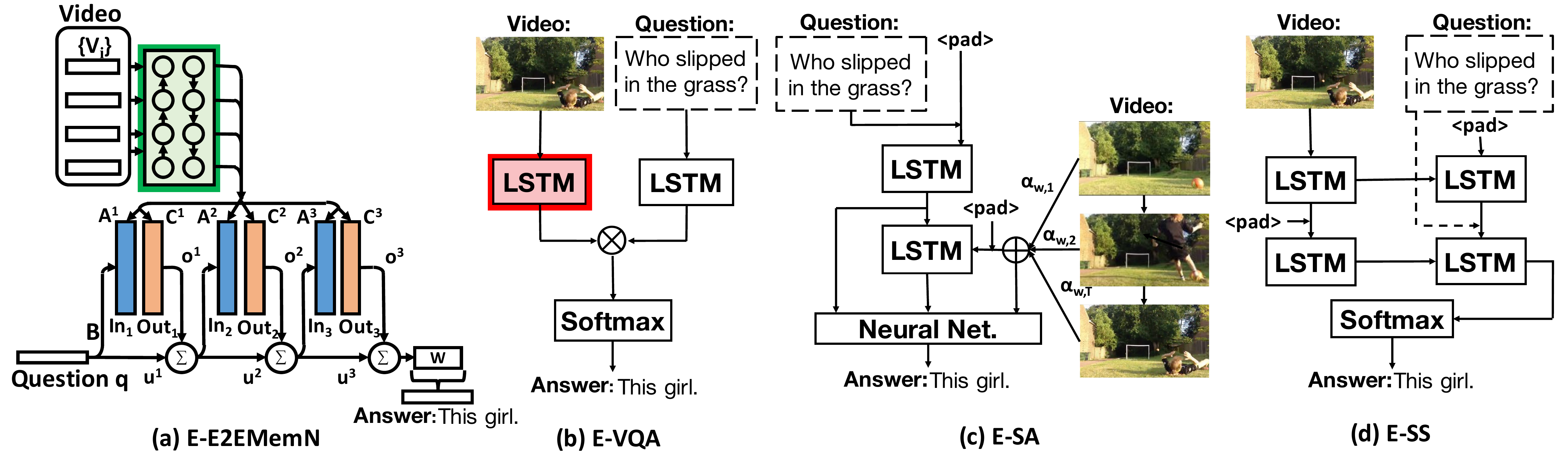}
    	\caption{\small Four extended methods for the video QA task. \textbf{(a)} The Extended End-to-End Memory Network (E-MN) uses additional bidirectional LSTM (green box) to encode the temporal information of videos. \textbf{(b)} The Extended VQA (E-VQA) model encodes temporal information using a single layer LSTM (red box). $\otimes$ denotes element-wise multiplication. \textbf{(c)} The Extended Soft-Attention (E-SA) model derives the question semantic meaning by LSTM encoding. $\oplus$ denotes element-wise addition. \textbf{(d)} The Extended Sequence-to-sequence (E-SS) model follows the video captioning system. Instead of decoding a caption, our model encodes a question and decodes an answer.}\label{fig:model}
    	\end{figure*}

\subsection{Extened Methods}\label{sec.VQAM}

We extend the following methods for our Video-QA task.

\noindent\textbf{Extended End-to-End Memory Network (MN)~\cite{NIPS2015_5846}.}
The QA task in MN consists of a set of statements, followed by a question whose answer is typically
a single word. We change the set of statements into a video -- a sequence of frames.
In order to capture the temporal relation among actions in consecutive frames, we first use a bi-directional LSTM to encode
the sequence of frame representations. The bi-directional LSTM and the MN are jointly
trained in an end-to-end fashion. Fig.~\ref{fig:model}(a) shows the model visualization similar to the one in~\cite{NIPS2015_5846}.

\noindent\textbf{Extended VQA~\cite{VQA}.}
The VQA model is designed for question answering given a single image observation.
We extend the model to handle video observation using a one-layer LSTM to encode a sequence of frames.
The extended E-VQA (Fig.~\ref{fig:model}(b)) encodes both video and question using two LSTMs separately into a joint representation space, where an AND-like operation (i.e., element-wise multiplication) is used to fuse two representations.

\noindent\textbf{Extended Soft Attention (SA)~\cite{VideoAttRNN}.}
The SA model learns to dynamically apply soft-attention on different frames in order to generate a caption.
We modified E-SA to encode questions while paying attention on different frames to generate an answer.
This model (Fig.~\ref{fig:model}(c)) mimics how humans understand a question while paying attention to different frames; finally, answer the question.

\noindent\textbf{Extended Sequence-to-sequence (SS)~\cite{SS15}.}
The SS model learns to encode a video; then, decode a sentence.
We modified E-SS to first, encode a video; then, encode a question; finally, decode an answer.
This model (Fig.~\ref{fig:model}(d)) mimics how humans first watch a video; then, listen to a question; finally, answer the question.

All extended QA methods consist of various combinations of sequence-encodings, embeddings, and soft-attention mechanisms.
They are all trained in an end-to-end fashion with our self-paced learning procedure outlined in the previous section.
We report their Video-QA performance in the experiments section.

\section{Experiments and Results}\label{sec.exp}

We evaluate all methods on our Video-QA dataset.
We use $14100$ videos and $151263$ candidate QA pairs for training,
$2000$ videos and $21352$ candidate QA pairs for validation,
and $2000$ videos and $2461$ ground truth QA pairs for testing.

\subsection{Implementation Details}

\noindent\textbf{QA pairs data preprocessing.} For simplicity, we do not explicitly stem, spellcheck or normalize any of the questions. We use a one-hot vector to represent words in the questions except for MN, where we use bag-of-words as in \cite{NIPS2015_5846}. We remove punctuations and replace digits for $<$NUMBER$>$. For answers, we only remove stop words. We choose the top K = $1000$ most frequent answers as possible candidates as in \cite{VQA}. This set covers $81\%$ of the training and validation answers.

\noindent\textbf{Video data preprocessing.} Similar to existing video understanding approaches, we utilize both appearance and local motion features. For appearance, we extract VGG~\cite{SimonyanICLR15} features for each frame. For local motion, we extract C3D~\cite{C3D15} features for $16$ consecutive frames.
We divide a video into maximum $45$-$50$ clips by considering GPU memory limit. Then, we average-pool all the VGG and C3D features in each clip to obtain a video observation $v$.

\noindent\textbf{Self-paced learning implementation.}
According to the results of data cleaning by Amazon Mechanical Turk, we found that about $10\%$ of the question-answer pairs are removed
by human annotators. Thus, at the first iteration of self-paced learning, we set $\gamma$ to remove $10\%$ QA pairs with small loss ratio in the training data. Then, the same $\gamma$ is used in all following iterations.
Our iterative self-paced method typically ends in $2$ iterations.

\begin{table*}[!htbp]
\centering
\caption{\small Video-QA results. We report the performance for \emph{Yes/No},  \emph{Others} questions, and their average (Avg.) separately in different columns. We evaluate the baseline method in the first row (\textbf{Baseline}), the four extended methods trained with all data in the second set of rows (\textbf{Train-all}), the four extended methods trained with no video observations in the third set of rows (\textbf{Non-visual}), and the four extended methods trained with self-paced learning in the last set of rows (\textbf{Self-paced}).
WUPS $0.0$ and WUPS $0.9$ are WUPS score thresholded by $0.0$ and $0.9$ respectively. $\textrm{Acc}^{\dagger}$ denotes classification accuracy penalizing false positive. Acc denotes classification accuracy. Avg. denotes the average classification accuracy of \emph{Others} and \emph{Yes/No}.}
\resizebox{0.97\textwidth}{!}{
\small
\begin{tabular}{|c||c||c|c|c|c||c|c|c|c||c|c|c|c|c|}
\hline
Video-QA & \multicolumn{1}{c|}{Baseline} & \multicolumn{4}{c|}{Train-all} &  \multicolumn{4}{c|}{Non-visual} &  \multicolumn{4}{c|}{Self-paced} \\ \hline
Others (\%) & ST & E-MN & E-VQA & E-SA & E-SS & E-MN & E-VQA & E-SA & E-SS & E-MN & E-VQA & E-SA & E-SS \\  \hline
WUPS 0.0  & 32.9 & 47.9  & 49.3 & 51.4 & 48.7 & 43.4 & \textbf{52.6} & 49.7 & 47.3 & 48.5 & 50.8 & 51.9 & 50.7  \\  \hline
WUPS 0.9 & 5.51 & 10.1 & 13.2 &15.5 & 14.2 & 8.2 & 10.0 & 12.4 & 11.2 &10.3 & 13.0 & \textbf{16.1} & 16.0 \\  \hline
Acc  & 2.1 & 2.9 & 5.0 & 8.4 & 7.3 & 1.8 & 2.1 & 4.7 & 4.8 & 3.0 & 5.1 & \textbf{9.4} & 9.3 \\  \hline \hline
Yes/No (\%) & ST & E-MN & E-VQA & E-SA & E-SS & E-MN & E-VQA & E-SA & E-SS & E-MN & E-VQA & E-SA & E-SS \\  \hline
Yes $\textrm{Acc}^{\dagger}$  & 11.9 & \textbf{40.0} & 38.8 & 36.4 & 34.5 & 35.4 & 39.0 & 38.9 & 39.5 & 30.4 & 39.1 & 39.0 & 39.7 \\  \hline
No $\textrm{Acc}^{\dagger}$  & 26.7 & 13.0 & 22.3 & \textbf{28.8} & 25.8 & 12.9 & 26.1 & 25.5 & 19.7 & 27.6 & 24.4 & 24.9 & 26.3 \\  \hline
Acc  & 49.3 & 49.5 & 46.7 & 52.4 & 49.5 & 50.0 & 48.3 & 51.6 & 49.6 & 52.0 & 47.8 & 51.6 & \textbf{52.7} \\  \hline \hline
Avg. (\%) & ST & E-MN & E-VQA & E-SA & E-SS & E-MN & E-VQA & E-SA & E-SS & E-MN & E-VQA & E-SA & E-SS \\  \hline
Acc  & 25.7 & 26.2 & 25.9 & 30.4 & 28.4 & 25.9 & 25.2 & 28.2 & 27.2 & 27.5 & 26.4 & 30.5 & \textbf{31.0} \\  \hline
\end{tabular}}
\label{table.QA}
\end{table*}

\begin{figure*}[!t]
\centering
\includegraphics[width=1.7\columnwidth]{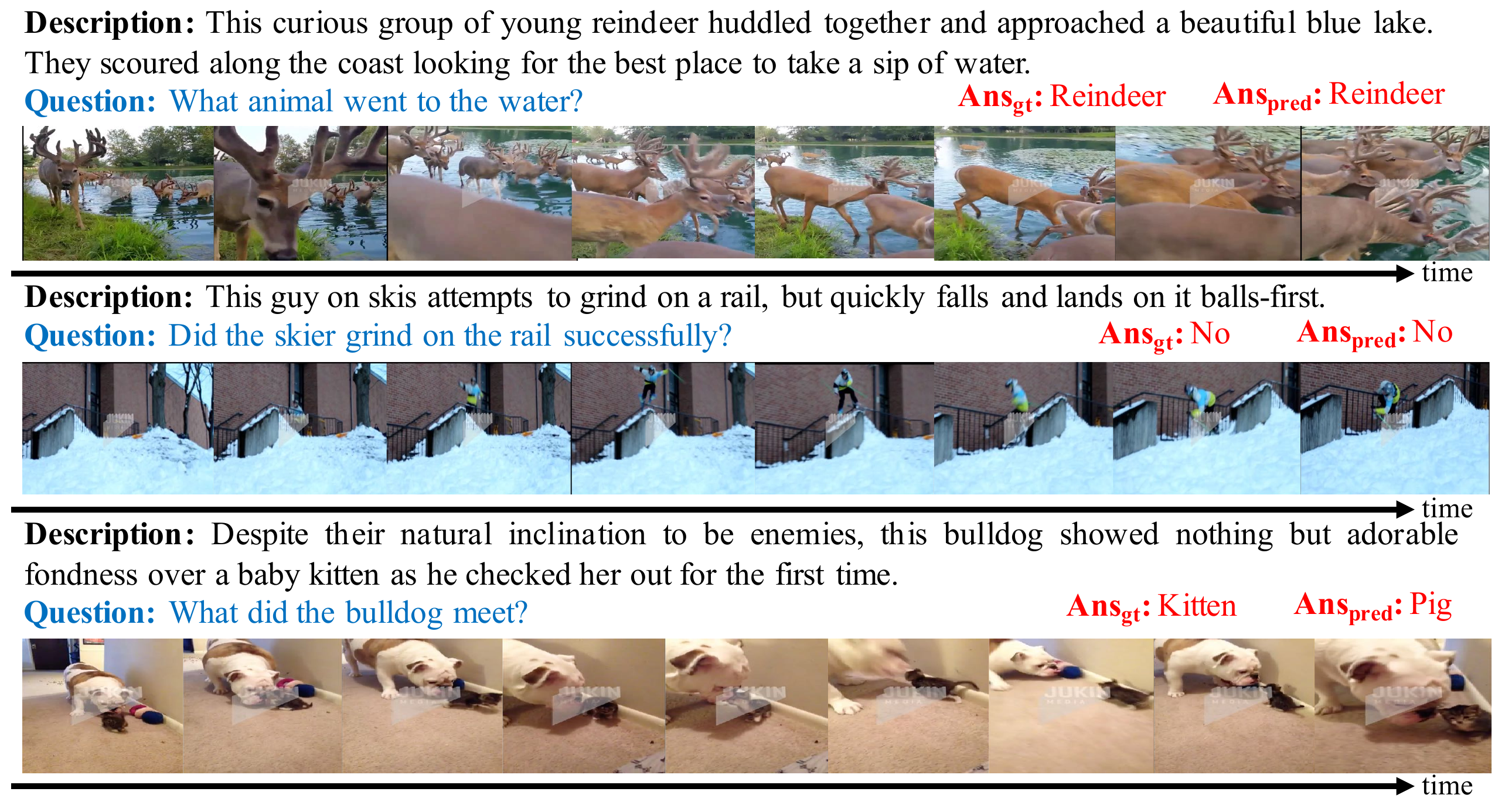}
\caption{\small Qualitative Video QA results. In each row, we show a typical examples of descriptions, questions, predicted answers and ground truth answers. The first and second one with corrected prediction are our good examples. The last one is a failure case, where the model is disctracted by the bulldog and mistakenly predicts it as a \emph{pig}. See \cite{VTGtech} for more examples.}\label{fig.typical}
\end{figure*}

\subsection{Training details}\label{sec.method_imp}

We implement and train all the extended methods using TensorFlow~\cite{tensorflow2015-whitepaper}
with the batch size of $100$ and selected the final model according to the best validation accuracy.
Other model-specific training details are described below.

\noindent\textbf{E-MN.} We use stochastic gradient descent with an initial learning rate of $0.001$, the same learning rate decay and gradient clipping scheme in~\cite{NIPS2015_5846}. Inspired by several memory based models, we set $500$ as the number of memories and the LSTM hidden dimension.

\noindent\textbf{E-VQA.} We use the same settings as in \cite{VQA}.

\noindent\textbf{E-SA.} We use the training settings as in \cite{VideoAttRNN}, except for
Adam optimization \cite{KingmaB14} with initial learning rate of $0.0001$.

\noindent\textbf{E-SS.} Except for the optimization algorithm and the total number of epochs, the training settings are all the same as ~\cite{SS15}. We use Adam optimizer~\cite{KingmaB14} with an initial learning rate of $0.0001$.

\subsection{Evaluation Metrics}\label{sec.eval_imp}

Inspired by Image-QA~\cite{NIPS2014_5411,VQA}, we evaluate Video-QA using both classification accuracy and the relaxed version of WUPS based on word similarity.
Notice that our answer space is $1K$ and classification accuracy is so strict that it will consider \emph{cat} a mistake when the ground truth is \emph{kitten}. Hence, we also report WUPS and use thresholds of $0.0$ and $0.9$ as in \cite{NIPS2014_5411}.
Moreover, we separately report performance on questions of the type \emph{Yes/No}  and \emph{Others},
as the former are considered to be less challenging. Finally, we report the average accuracy over \emph{Yes/No} and \emph{Others} (see Table.~\ref{table.QA}).

\subsection{Results}\label{sec.quant}

\noindent\textbf{Baseline method.}
We use Skip-Thought (ST) \cite{kiros2015skip} to directly learn the sentence semantic and syntactic properties in a Recurrent Neural Network framework.
Using the above methods as the representation for questions, we can capture the similarity between question sentences. 
Given a test question, we retrieve the top $10$ nearest (using cosine similarity) training questions and their answers. 
The final answer is chosen by the majority votes of the top ten answer list.
We compare the extended methods with the question retrieval baseline in the \textbf{Baseline} section of Table.~\ref{table.QA}.
We found that baseline performs significantly worse than our extended methods on \emph{Others} questions,
but performs on a par with extended methods on \emph{Yes/No} questions.
Hence, we suspect the baseline makes many false positive \emph{Yes/No} predictions.
For \emph{Yes/No}, we further report $\frac{\textrm{true-positive}}{\textrm{true-positive+false-positive+false-negative}}$ as $\textrm{Acc}^{\dagger}$,
which penalizes false positive predictions. As measured by $\textrm{Acc}^{\dagger}$, the baseline is inferior to most extended methods.

\noindent\textbf{Extended methods.}
Self-paced E-SS ($31.0\%$ average Acc) outperforms other extended methods since it jointly encodes both videos and questions sequentially.
On the other hand,  self-paced E-VQA performs the worst among all extended methods, since it only uses an element-wise multiplication operation to combine visual observation and questions.

\noindent\textbf{Importance of video observation.}
We also train all extended methods with dummy video observations such that they are forced to answer the only given question.
In the \textbf{Non-visual} section of Table.~\ref{table.QA}, we show that all extended methods suffer when not observing videos.

\noindent\textbf{Effectiveness of self-paced learning.}
In the \textbf{Self-paced} section of Table.~\ref{table.QA}, we show that all extended methods achieve performance gain after self-paced learning.
E-SA achieves the smallest gain since soft-attention (SA) can select different visual observations to handle noisy QA training pairs.
Among them, {E-SS} achieves a $2.6\%$ improvement in average accuracy over its \textbf{Train-all} version.
Finally, we show typical Video-QA results of our best method (E-SS) in Fig.~\ref{fig.typical} and more examples in technical report \cite{VTGtech}.


\section{Conclusions}\label{sec.conclusions}
Our scalable approach has generated a large-scale video-based question answering dataset (e.g., $18100$ videos and $175076$ QA pairs) with minimal human effort.
Moreover, our extended models and self-paced learning procedure
are shown to be effective.
In the future, we will further increase the scale of the Video-QA dataset
and improve the procedure to handle a larger amount of non-perfect training examples.

\section{Acknowledgments}
We thank Microsoft Research Asia, MOST 103-2218-E-007-022, MOST 104-3115-E-007-005, MOST 105-2218-E-007-012, NOVATEK Fellowship, and Panasonic for their support. We also thank Shih-Han Chou, Heng Hsu, and I-Hsin Lee for their collaboration with the dataset.

\bibliographystyle{aaai}

\begin{thebibliography}{}

\bibitem[\protect\citeauthoryear{Andreas \bgroup et al\mbox.\egroup
  }{2016}]{AndreasCVPR16}
Andreas, J.; Rohrbach, M.; Darrell, T.; and Klein, D.
\newblock 2016.
\newblock Deep compositional question answering with neural module networks.
\newblock In {\em CVPR}.

\bibitem[\protect\citeauthoryear{Antol \bgroup et al\mbox.\egroup }{2015}]{VQA}
Antol, S.; Agrawal, A.; Lu, J.; Mitchell, M.; Batra, D.; Zitnick, C.~L.; and
  Parikh, D.
\newblock 2015.
\newblock {VQA}: Visual question answering.
\newblock In {\em ICCV}.

\bibitem[\protect\citeauthoryear{Bigham \bgroup et al\mbox.\egroup
  }{2010}]{Bigham:2010:VNR}
Bigham, J.~P.; Jayant, C.; Ji, H.; Little, G.; Miller, A.; Miller, R.~C.;
  Miller, R.; Tatarowicz, A.; White, B.; White, S.; and Yeh, T.
\newblock 2010.
\newblock Vizwiz: Nearly real-time answers to visual questions.
\newblock In {\em UIST}.

\bibitem[\protect\citeauthoryear{et al.}{2015}]{tensorflow2015-whitepaper}
et~al., M.~A.
\newblock 2015.
\newblock {TensorFlow}: Large-scale machine learning on heterogeneous systems.
\newblock Software available from tensorflow.org.

\bibitem[\protect\citeauthoryear{Gao \bgroup et al\mbox.\egroup
  }{2015}]{gao2015mQA}
Gao, H.; Mao, J.; Zhou, J.; Huang, Z.; Wang, L.; and Xu, W.
\newblock 2015.
\newblock Are you talking to a machine? dataset and methods for multilingual
  image question answering.
\newblock In {\em NIPS}.

\bibitem[\protect\citeauthoryear{Gates}{2008}]{GatesThesis08}
Gates, D.~M.
\newblock 2008.
\newblock Generating reading comprehension look-back strategy questions from
  expository texts.
\newblock Master's thesis, Carnegie Mellon University.

\bibitem[\protect\citeauthoryear{Geman \bgroup et al\mbox.\egroup
  }{2014}]{Geman24032015}
Geman, D.; Geman, S.; Hallonquist, N.; and Younes, L.
\newblock 2014.
\newblock Visual turing test for computer vision systems.
\newblock {\em PNAS} 112(12):3618--3623.

\bibitem[\protect\citeauthoryear{Heilman and Smith}{2010}]{Heilman2010}
Heilman, M., and Smith, N.~A.
\newblock 2010.
\newblock Good question! statistical ranking for question generation.
\newblock In {\em HLT}.

\bibitem[\protect\citeauthoryear{Hendricks \bgroup et al\mbox.\egroup
  }{2016}]{hendricks2015deep}
Hendricks, L.~A.; Venugopalan, S.; Rohrbach, M.; Mooney, R.; Saenko, K.; and
  Darrell, T.
\newblock 2016.
\newblock Deep compositional captioning: Describing novel object categories
  without paired training data.
\newblock {\em CVPR}.

\bibitem[\protect\citeauthoryear{Kingma and Ba}{2015}]{KingmaB14}
Kingma, D.~P., and Ba, J.
\newblock 2015.
\newblock Adam: {A} method for stochastic optimization.
\newblock {\em ICLR}.

\bibitem[\protect\citeauthoryear{Kiros \bgroup et al\mbox.\egroup
  }{2015}]{kiros2015skip}
Kiros, R.; Zhu, Y.; Salakhutdinov, R.~R.; Zemel, R.; Urtasun, R.; Torralba, A.;
  and Fidler, S.
\newblock 2015.
\newblock Skip-thought vectors.
\newblock In {\em NIPS}.

\bibitem[\protect\citeauthoryear{Ma, Lu, and Li}{2016}]{MaAAAI16}
Ma, L.; Lu, Z.; and Li, H.
\newblock 2016.
\newblock Learning to answer questions from image using convolutional neural
  network.
\newblock In {\em AAAI}.

\bibitem[\protect\citeauthoryear{Malinowski and Fritz}{2014}]{NIPS2014_5411}
Malinowski, M., and Fritz, M.
\newblock 2014.
\newblock A multi-world approach to question answering about real-world scenes
  based on uncertain input.
\newblock In {\em NIPS}.

\bibitem[\protect\citeauthoryear{Malinowski, Rohrbach, and
  Fritz}{2015}]{Malinowski_2015_ICCV}
Malinowski, M.; Rohrbach, M.; and Fritz, M.
\newblock 2015.
\newblock Ask your neurons: A neural-based approach to answering questions
  about images.
\newblock In {\em ICCV}.

\bibitem[\protect\citeauthoryear{Noh, Seo, and Han}{2016}]{noh2015dppnet}
Noh, H.; Seo, P.~H.; and Han, B.
\newblock 2016.
\newblock Image question answering using convolutional neural network with
  dynamic parameter prediction.
\newblock {\em CVPR}.

\bibitem[\protect\citeauthoryear{Pan \bgroup et al\mbox.\egroup
  }{2016a}]{PanXYWZ15}
Pan, P.; Xu, Z.; Yang, Y.; Wu, F.; and Zhuang, Y.
\newblock 2016a.
\newblock Hierarchical recurrent neural encoder for video representation with
  application to captioning.
\newblock {\em CVPR}.

\bibitem[\protect\citeauthoryear{Pan \bgroup et al\mbox.\egroup
  }{2016b}]{PanMYLR15}
Pan, Y.; Mei, T.; Yao, T.; Li, H.; and Rui, Y.
\newblock 2016b.
\newblock Jointly modeling embedding and translation to bridge video and
  language.
\newblock {\em CVPR}.

\bibitem[\protect\citeauthoryear{Ren, Kiros, and Zemel}{2015}]{NIPS2015Ren}
Ren, M.; Kiros, R.; and Zemel, R.
\newblock 2015.
\newblock Exploring models and data for image question answering.
\newblock In {\em NIPS}.

\bibitem[\protect\citeauthoryear{Rus and Graessar}{2009}]{QGChallenge09}
Rus, V., and Graessar.
\newblock 2009.
\newblock Question generation shared task and evaluation challenge ¡v status
  report.
\newblock In {\em The Question Generation Shared Task and Evaluation
  Challenge}.

\bibitem[\protect\citeauthoryear{Rus and Lester}{2009}]{QGworkshop09}
Rus, V., and Lester, J.
\newblock 2009.
\newblock Workshop on question generation.
\newblock In {\em Workshop on Question Generation}.

\bibitem[\protect\citeauthoryear{Simonyan and Zisserman}{2015}]{SimonyanICLR15}
Simonyan, K., and Zisserman, A.
\newblock 2015.
\newblock Very deep convolutional networks for large-scale image recognition.
\newblock In {\em ICLR}.

\bibitem[\protect\citeauthoryear{Sukhbaatar \bgroup et al\mbox.\egroup
  }{2015}]{NIPS2015_5846}
Sukhbaatar, S.; szlam, a.; Weston, J.; and Fergus, R.
\newblock 2015.
\newblock End-to-end memory networks.
\newblock In {\em NIPS}.

\bibitem[\protect\citeauthoryear{Tapaswi \bgroup et al\mbox.\egroup
  }{2016}]{MovieQA}
Tapaswi, M.; Zhu, Y.; Stiefelhagen, R.; Torralba, A.; Urtasun, R.; and Fidler,
  S.
\newblock 2016.
\newblock {MovieQA}: Understanding stories in movies through
  question-answering.
\newblock In {\em CVPR}.

\bibitem[\protect\citeauthoryear{Tran \bgroup et al\mbox.\egroup
  }{2015}]{C3D15}
Tran, D.; Bourdev, L.; Fergus, R.; Torresani, L.; and Paluri, M.
\newblock 2015.
\newblock Learning spatiotemporal features with 3d convolutional networks.
\newblock In {\em ICCV}.

\bibitem[\protect\citeauthoryear{Tu \bgroup et al\mbox.\egroup
  }{2014}]{TuMLCZ14}
Tu, K.; Meng, M.; Lee, M.~W.; Choe, T.~E.; and Zhu, S.~C.
\newblock 2014.
\newblock Joint video and text parsing for understanding events and answering
  queries.
\newblock In {\em IEEE MultiMedia}.

\bibitem[\protect\citeauthoryear{Venugopalan \bgroup et al\mbox.\egroup
  }{2015}]{SS15}
Venugopalan, S.; Rohrbach, M.; Donahue, J.; Mooney, R.; Darrell, T.; and
  Saenko, K.
\newblock 2015.
\newblock Sequence to sequence - video to text.
\newblock In {\em ICCV}.

\bibitem[\protect\citeauthoryear{Vinyals \bgroup et al\mbox.\egroup
  }{2015}]{showtell2015}
Vinyals, O.; Toshev, A.; Bengio, S.; and Erhan, D.
\newblock 2015.
\newblock Show and tell: A neural image caption generator.
\newblock In {\em CVPR}.

\bibitem[\protect\citeauthoryear{Xu \bgroup et al\mbox.\egroup
  }{2016}]{xu2016msr}
Xu, J.; Mei, T.; Yao, T.; and Rui, Y.
\newblock 2016.
\newblock Msr-vtt: A large video description dataset for bridging video and
  language.
\newblock In {\em Conference on Computer Vision and Pattern Recognition
  (CVPR)}.

\bibitem[\protect\citeauthoryear{Yao \bgroup et al\mbox.\egroup
  }{2015}]{VideoAttRNN}
Yao, L.; Torabi, A.; Cho, K.; Ballas, N.; Pal, C.; Larochelle, H.; and
  Courville., A.
\newblock 2015.
\newblock Describing videos by exploiting temporal structure.
\newblock In {\em ICCV}.

\bibitem[\protect\citeauthoryear{Yu \bgroup et al\mbox.\egroup
  }{2016}]{YuWHYX15}
Yu, H.; Wang, J.; Huang, Z.; Yang, Y.; and Xu, W.
\newblock 2016.
\newblock Video paragraph captioning using hierarchical recurrent neural
  networks.
\newblock {\em CVPR}.

\bibitem[\protect\citeauthoryear{Zeng \bgroup et al\mbox.\egroup
  }{2016a}]{VTGtech}
Zeng, K.-H.; Chen, T.-H.; Chuang, C.-Y.; Liao, Y.-H.; Niebles, J.~C.; and Sun,
  M.
\newblock 2016a.
\newblock Technical report: Leveraging video descriptions to learn video
  question answering.
\newblock \url{http://aliensunmin.github.io/project/video-language/}.

\bibitem[\protect\citeauthoryear{Zeng \bgroup et al\mbox.\egroup
  }{2016b}]{title_generation}
Zeng, K.-H.; Chen, T.-H.; Niebles, J.~C.; and Sun, M.
\newblock 2016b.
\newblock Title generation for user generated videos.
\newblock In {\em ECCV}.

\bibitem[\protect\citeauthoryear{Zhu \bgroup et al\mbox.\egroup
  }{2015}]{zhu2015uncovering}
Zhu, L.; Xu, Z.; Yang, Y.; and Hauptmann, A.~G.
\newblock 2015.
\newblock Uncovering temporal context for video question and answering.
\newblock {\em arXiv preprint arXiv:1511.04670}.

\end{thebibliography}

\end{document}